# Domain Adaptation for Deviating Acquisition Protocols in CNN-based Lesion Classification on Diffusion-Weighted MR Images

Jennifer Kamphenkel⋆[1], Paul F. Jäger⋆[1], Sebastian Bickelhaupt[2], Frederik Bernd Laun[2,3], Wolfgang Lederer[4], Heidi Daniel[5], Tristan Anselm Kuder[6], Stefan Delorme[2], Heinz-Peter Schlemmer[2], Franziska König[2], and Klaus H. Maier-Hein[1]

[1] Division of Medical Image Computing, German Cancer Research Center (DKFZ), Heidelberg, Germany
[2] Department of Radiology, DKFZ, Heidelberg, Germany
[3] Institute of Radiology, University Hospital Erlangen, Germany
[4] Radiological Practice at the ATOS Clinic, Heidelberg, Germany
[5] Radiology Center Mannheim (RZM), Germany
[6] Medical Physics in Radiology, DKFZ, Heidelberg, Germany

**Abstract.** End-to-end deep learning improves breast cancer classification on diffusion-weighted MR images (DWI) using a convolutional neural network (CNN) architecture. A limitation of CNN as opposed to previous model-based approaches is the dependence on specific DWI input channels used during training. However, in the context of large-scale application, methods agnostic towards heterogeneous inputs are desirable, due to the high deviation of scanning protocols between clinical sites. We propose model-based domain adaptation to overcome input dependencies and avoid re-training of networks at clinical sites by restoring training inputs from altered input channels given during deployment. We demonstrate the method's significant increase in classification performance and superiority over implicit domain adaptation provided by training-schemes operating on model-parameters instead of raw DWI images.

## 1 Introduction

As mammography suffers from high amounts of false positive findings, a promising image modality for breast cancer classification is DWI, which aims at reducing the number of biopsies through reliable early diagnosis [1]. The model-based state of the art for DWI signal exploitation is diffusion kurtosis imaging (DKI), where diffusion properties are estimated in suspicious tissue to distinguish between malignant and benign tumor cells [2,3]. An end-to-end q-space deep learning approach (E2E) has recently been shown to outperform DKI-based approaches by optimally exploiting input correlations using CNNs [4,5]. However, a limitation of E2E is the inherent input dependence of CNNs [6], which in this case are trained on specific diffusion-weighted images acquired at certain *b-values*, i.e. strengths and timings of gradient fields. This limitation is crucial for large-scale clinical application, since DWI scanning protocols deviate between sites and standardization is not expected in the near future. Furthermore, due to limited training data, it is desirable to ship trained models across clinical sites for inference on unseen images acquired with arbitrary local protocols. This procedure implies heterogeneities between training data and local inference data, e.g. in the form of *shifted* or *missing* b-values.

⋆ contributed equally

Generative models such as generative adversarial networks [7,8] and variational autoencoders [9,10] have recently succeeded at domain transformations. Such models could potentially be used to transform altered test-time inputs to original input channels used during training, yet do not eliminate input dependencies. Similar to other domain adaptation methods such as fine-tuning of models on new input or common representation learning of inputs [11], they themselves need to be trained on specific input alteration modes. As model fits such as DKI come with an inherent robustness towards input variations, input independence could potentially be achieved by operating on the fit parameters instead of raw DWI inputs. However, this robustness is proportional to the number of observed values, which, as will be shown, is not sufficient in typical DWI acquisition setups.

In this paper, we propose model-based domain adaptation, where the original training channels are derived from DKI using the altered inputs at test time. This method does not require training and hence can be deployed in any clinical setting without prior assumptions about protocol deviations. We show that this method significantly reduces input dependencies by optimally exploiting input correlations (E2E) based on estimations from the DKI model. We further demonstrate the superiority of our approach over training networks on DKI parameters (fit-to-end, F2E).

## 2 Methods

### 2.1 DWI Data Set

This study is performed on a data set of 221 patients and is equal to the data set used for E2E training [4,5]. For each patient, images of four b-values 0, 100, 750 and 1500 s $mm^{-2}$ with a slice thickness of 3 mm were acquired using two different 1.5 T MR scanners. The in-plane resolution of one scanner had to be upsampled by a factor 2 to match the other scanners resolution of 1.25 mm. Prior to DWI scanning, all patients were diagnosed with BI-RADS [12] $\geq$4 from mammography screenings. A core-needle biopsy was performed to secure diagnosis, which resulted in 121 malignant and 100 benign lesions. The biopsy result served as the classification ground truth. Lesions were manually segmented as regions of interest (ROI) by expert radiologist without knowledge about the biopsy results. As 23 images do not contain any visible lesion, those subjects were predicted as benign. Figure 1 shows an example set of diffusion-weighted images for one patient.

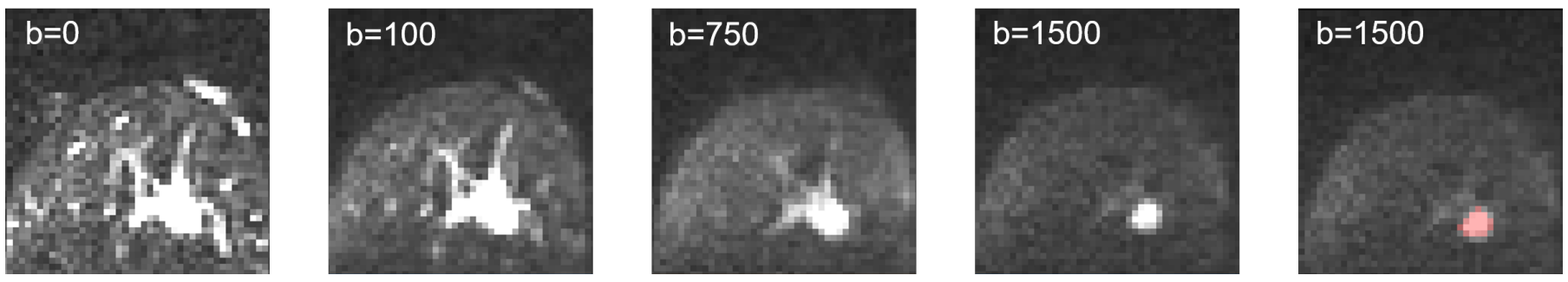

**Fig. 1:** Sample slice of diffusion-weighted images of one patient at distinct b-values and the segmentation of the lesion on b = 1500 s $mm^{-2}$ (right).

### 2.2 Diffusion Kurtosis Imaging

DKI is the the state of the art model for DWI signal exploitation in lesion classification. To derive diagnostically conclusive tissue parameters, DKI estimates the apparent diffusion coefficient (ADC) and additionally the apparent kurtosis coefficient (AKC) which quantifies deviations from free Gaussian diffusion induced by diffusion restrictions and diffusion heterogeneity [13]. These parameters are estimated by fitting the DKI model to measured signal intensities $S(b)$ in each voxel:

$$S(b) = (\theta^2 + S_0 \ exp(-b \ ADC + \frac{1}{6} \ b^2 \ ADC^2 \ AKC)^2)^{0.5} \tag{1}$$

where $S_0$ is the signal intensity for b0 (b = 0), the b-value is the strength of diffusion weighting [14]. Furthermore, the model accounts for a background signal level induced by fat signal contamination in the lesion using the mean signal intensity $\theta$ of an additionally segmented fat area for each patient. In DKI, ADC and AKC are used most commonly to determine the malignancy of a suspicious lesion by averaging the coefficients over an ROI to obtain global coefficients [2]. Notably, we updated the DKI fit of [5] by not omitting $S(0)$ and added fat calibration to increase DKI fitting performance according to [14].

### 2.3 End-to-end q-space Deep Learning

E2E has recently been proposed as a successful model-free approach to classifying suspicious breast lesions [4,5]. Classification is performed by feeding the raw signal intensities of the segmented ROI into a CNN. Using 1x1 convolutions, deep diffusion coefficients are learned mimicking DKI parameters by correlating signal intensities of each pixel across DWI input channels. Subsequently, the network extracts features related to texture and geometry, which are globally pooled and fed through a softmax layer to obtain probabilities of malignancy.

### 2.4 Model-based Domain Adaptation

To overcome dependence on specific b-values and enable clinical applicability of lesion classification regardless of scanning protocols, we propose to perform model-based domain adaptation (MBDA). During inference, the DKI model is fit to the signal intensities of all available (potentially altered) b-values. In order to restore the original set of b-values seen during training, the fitted model is used to derive estimates of the signal intensities $S(b)$ at the missing b-values (see Formula 1). Subsequently, the restored set of inputs is fed into the trained model to obtain classification scores (see Figure 2 top).

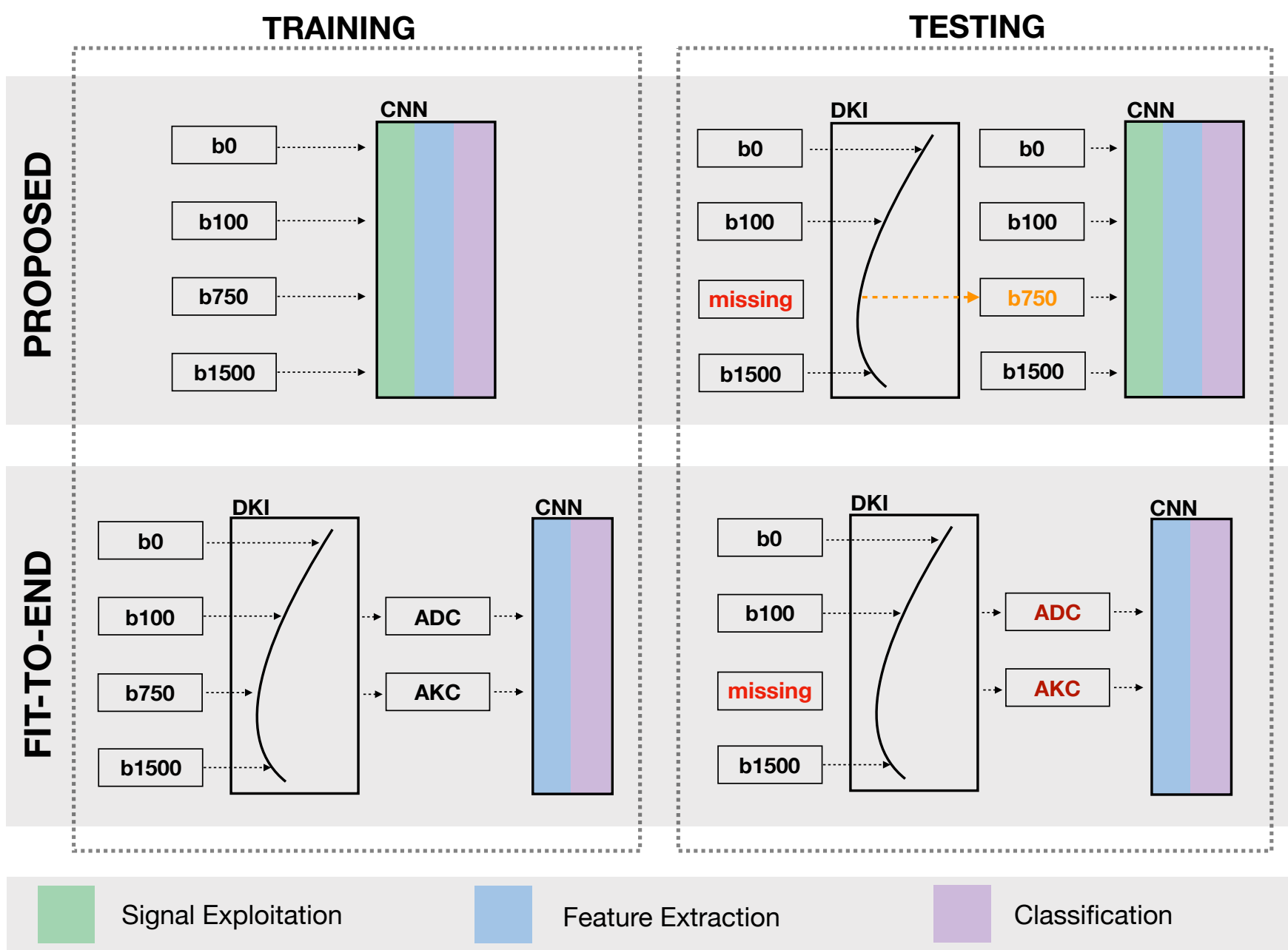

**Fig. 2:** Concept of our proposed method for the *missing scenario* (top). The missing b-value is derived from a DKI-model and used as CNN input. The fit-to-end architecture trained on ADC and AKC is used for comparison (bottom).

**Experimental Setup** Two scenarios of heterogeneous inputs were studied: *shifted scenario*, where one measured b-value in the inference data is provided at a different (shifted) value w.r.t. the training data, and *missing scenario*, where one measured b-value in the inference data is missing w.r.t. the training data. Both scenarios were imitated by training and testing on respective subsets of the four b-values provided by the utilized data set. Note, that scenarios comprising alterations of multiple inputs were not studied due to the limited number of b-values provided. Furthermore, no alterations were applied to b0 as in practice all protocols include at least one b-value equal or close to zero [13,14,15].

An upper bound performance for MBDA is given by training and testing on the same subset of b-values (*matched input*). A lower bound performance for MBDA is given by testing on the altered inputs without domain adaptation (*altered input*). To compare our approach against the implicit domain adaptation of DKI, we train on DKI fit parameters ADC and AKC by feeding the parameter maps directly into the feature extraction and classification modules of the CNN (F2E). During testing, ADC and AKC are fitted using the altered inputs (see Figure 2 bottom). For inference subsets containing only two b-values, which causes the DKI model to be under-constrained, we set $AKC = 0$.

The network details and training setup are equal to the setup reported in [5]. The signal exploitation module is omitted for F2E training. The networks are trained using 5-fold cross validation with with 60% training- , 20% validation- and 20% test data and selected based on the lowest validation error.

**Evaluation** Evaluation is conducted by comparing the area under the receiver operator curves (AUC). Significance tests were performed using DeLong's method and corrected for multiple testing using the Holm-Bonferroni-Method (initial $\alpha = 0.05$).

## 3 Results

Results are shown in Table 1. The observed moderate decrease of performance caused by a general absence of inputs (*matched input*) indicates a general redundancy of information across b-values of the input images. For instance, subsets of three b-values seem to roughly contain the same information as the original four b-values with respect to overall performance. However, strong input dependence is observed in both E2E and F2E (*altered input*, i.e. no domain adaptation) with an average decrease of 19.2% and 10.6%. MBDA is able to significantly increase this lower bound performance in the shifted scenario (12.4%) and missing scenario (16.8%) (see Figure 3). Comparing F2E to E2E, F2E *altered input* performs on average slightly better than E2E *altered input*, i.e. 7.1% for shifted scenario and 4.4% for missing scenario, indicating a positive effect of implicit domain adaptation. E2E with MDBA considerably outperforms F2E by 5.3% for shifted scenario and 12.4% for missing scenario. Notably, extrapolation to large b-values is a poorly constrained problem, which causes performance drops across all explored methods. As expected, F2E only works when constraining the DKI model (setting $AKC = 0$) during CNN training.

**Table 1:** Results comparing all explored methods. All numbers report AUC except for p-values. *x* marks the available b-values. *o* marks the derived b-value. * marks observed significance.

**a) Shifted Scenario.**

| Training b-values | | | | E2E Matched Input | F2E Matched Input | Testing b-values | | | | E2E Altered Input | F2E Altered Input | MBDA | p-value E2E;MBDA | p-value E2E;F2E |
|---|---|---|---|---|---|---|---|---|---|---|---|---|---|---|
| b0 | b100 | b750 | b1500 | | | b0 | b100 | b750 | b1500 | | | | | |
| x | x | x | | 0.893±0.04 | 0.819±0.05 | x | x | o | x | 0.741±0.06 | 0.768±0.05 | **0.848**±0.05 | 0.0005* | 0.011 |
| | | | | | | x | o | x | x | 0.831±0.05 | 0.845±0.05 | **0.893**±0.04 | 0.0052* | 0.0622 |
| x | x | | x | 0.882±0.04 | 0.855±0.05 | x | x | x | o | 0.799±0.06 | **0.817**±0.06 | 0.751±0.07 | 0.1426 | 0.1132 |
| | | | | | | x | o | x | x | 0.831±0.05 | 0.845±0.05 | **0.880**±0.04 | 0.0019* | 0.816 |
| x | | x | x | 0.886±0.04 | 0.892±0.04 | x | x | x | o | 0.725±0.07 | **0.845**±0.05 | 0.766±0.07 | 0.3199 | 0.0416 |
| | | | | | | x | x | o | x | 0.737±0.07 | 0.844±0.05 | **0.871**±0.05 | 6.96e-5* | 0.422 |
| x | x | | | 0.777±0.06 | 0.674±0.072 | x | o | x | | 0.680±0.07 | 0.679±0.07 | **0.794**±0.06 | 0.00014* | 0.0018* |
| | | | | | | x | o | | x | 0.666±0.07 | 0.679±0.07 | **0.791**±0.06 | 0.0002* | 0.0015* |
| x | | x | | 0.889±0.04 | 0.871±0.05 | x | x | o | | 0.723±0.07 | 0.608±0.08 | **0.796**±0.06 | 0.0467 | 4.08e-6* |
| | | | | | | x | | o | x | 0.752±0.06 | 0.833±0.06 | **0.869**±0.05 | 0.0009* | 0.1426 |
| x | | | x | 0.882±0.04 | 0.877±0.05 | x | x | | o | 0.729±0.07 | 0.589±0.08 | **0.757**±0.06 | 0.4864 | 0.0002* |
| | | | | | | x | | x | o | 0.817±0.06 | 0.825±0.06 | **0.866**±0.05 | 0.0643 | 0.1485 |

**b) Missing Scenario.**

As for subsets of two available b-value images DKI is manually constrained by setting $AKC = 0$, performances for both training with and without the constraint are reported (DKI/ADC)

| Training b-values | | | | E2E Matched Input | F2E Matched Input (DKI/ADC) | Testing b-values | | | | E2E Altered Input | F2E Altered Input | MBDA | p-value E2E;MBDA | p-value E2E;F2E (DKI/ADC) |
|---|---|---|---|---|---|---|---|---|---|---|---|---|---|---|
| b0 | b100 | b750 | b1500 | | | b0 | b100 | b750 | b1500 | | | | | |
| x | x | x | x | 0.898±0.05 | 0.896±0.05 | x | x | x | o | 0.678±0.07 | 0.655±0.07 | **0.745**±0.07 | 0.1463 | 0.0449* |
| | | | | | | x | x | o | x | 0.604±0.08 | 0.667±0.07 | **0.882**±0.04 | 1.4e-12* | 8.76e-8* |
| | | | | | | x | o | x | x | 0.823±0.53 | 0.678±0.07 | **0.901**±0.04 | 0.00028* | 1.04e-8* |
| x | x | x | | 0.893±0.04 | 0.819±0.05/ 0.859±0.05 | x | x | o | | 0.513±0.08 | 0.522±0.08/ 0.617±0.07 | **0.780**±0.06 | 2.1e-7* | 1.18e-8*/ 0.00014* |
| | | | | | | x | o | x | | 0.817±0.05 | 0.514±0.08/ 0.857±0.08 | **0.891**±0.04 | 0.00026* | 2.2e-16*/ 0.1041 |
| x | x | | x | 0.882±0.04 | 0.855±0.05/ 0.860±0.05 | x | x | | o | 0.512±0.08 | 0.612±0.08/ 0.652±0.074 | **0.755**±0.06 | 6.92e-6* | 0.00067*/ 0.0125* |
| | | | | | | x | o | | x | 0.818±0.05 | 0.647±0.08/ 0.875±0.05 | **0.879**±0.04 | 0.0003* | 3.63e-9*/ 0.8804 |
| x | | x | x | 0.886±0.04 | 0.892±0.04/ 0.860±0.05 | x | | x | o | 0.657±0.07 | 0.646±0.07/ 0.836±0.05 | **0.878**±0.04 | 5.14e-9* | 8.72e-10*/ 0.1036 |
| | | | | | | x | | o | x | 0.649±0.07 | 0.699±0.07/ **0.868**±0.05 | **0.868**±0.04 | 3.24e-7* | 2.66e-6*/ 0.997 |

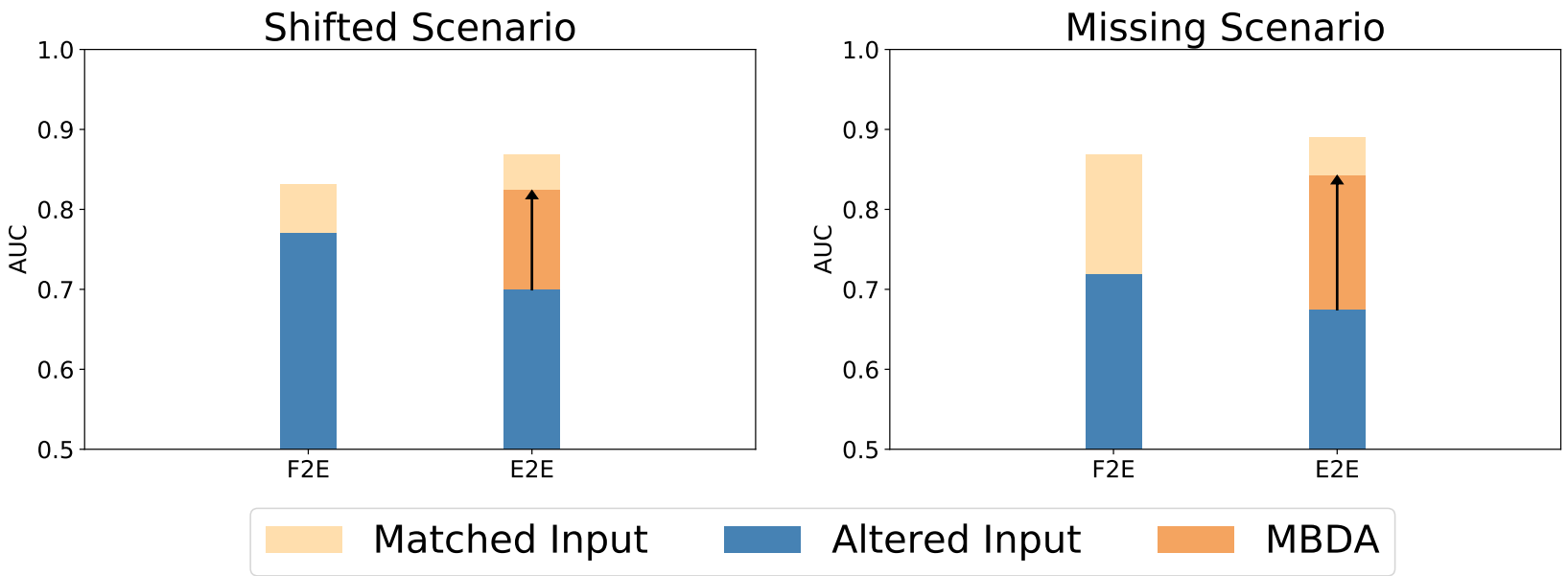

**Fig. 3:** Mean AUC derived from Table 1. Matched input represents the upper bound with matching b-value subsets during training and inference. Altered Input represents the lower bound by testing on the altered subset without domain adaptation. E2E with MBDA significantly improves the robustness towards heterogeneous inputs compared to F2E with altered inputs (implicit domain adaptation) in both scenarios.

## 4 Discussion

The results of this study suggest that model-based domain adaptation is an effective approach to overcome input dependencies and avoid re-training at clinical sites during large-scale application of DWI lesion classification. MBDA significantly increases the performance for both missing and shifted input scenarios by combining optimal exploitation of input correlations of raw DWI with DKI-based signal estimation to restore information lost due to altered input. In other words, MBDA is a “minimal invasive” method, which leaves unaltered input untouched, while the implicit domain adaptation performed by training and testing on fit parameters generates entirely new fit parameters given altered input, discarding unaltered correspondences. The latter works in theory, given a sufficient number of b-value images, but suffers from fitting instabilities in a typical DWI setup. In addition, strong assumptions have to be made on the amount of b-value images available during clinical inference prior to CNN training (as manually constraining the model by setting $AKC = 0$ might be required), which contradicts the desire for input independence. Future research includes studying multiple input alterations on data sets providing a larger number of b-values, application on unsegmented breast DWI, investigating the generalization of deep learning models trained on large DWI data sets and exploring the applicability to further entities.